\RequirePackage{fix-cm}
\documentclass[twocolumn]{svjour3}          
\smartqed  
\usepackage{graphicx}
\usepackage{amsmath,amssymb} 
\usepackage[utf8x]{inputenc}
\usepackage{pgfplots}
\pgfplotsset{compat=newest}
\usepackage{tikz}
\usetikzlibrary{backgrounds}
\usepackage{color}
\usepackage{xcolor}
\usepackage{booktabs,siunitx} 
\usepackage{makecell}
\usepackage[noadjust]{cite}
\usepackage[colorlinks=true,allcolors=blue]{hyperref}
\usepackage{rotating}

\usepackage{times}
\usepackage{pifont}
\usepackage{arydshln}
\usepackage{adjustbox}
\usepackage{array,multirow}
\usepackage{subfigure}
\usepackage{paralist}
\usepackage[english]{babel}

\usepackage{booktabs}
\usepackage{threeparttable}

\usepackage{authblk}
\usepackage{diagbox}

\usepackage{natbib}
\setcitestyle{authoryear,open={(},close={)}}

\begin{document}

\title{UCL-Dehaze: Towards Real-world Image Dehazing via Unsupervised Contrastive Learning}

\titlerunning{UCL-Dehaze: Towards Real-world Image Dehazing via Unsupervised Contrastive Learning}

\author{Yongzhen Wang         \and
        Xuefeng Yan \and
        Fu Lee Wang \and
        Haoran Xie \and
        Wenhan Yang \and
        Mingqiang Wei \and
        Jing Qin 
}

\institute{Yongzhen Wang  \and Xuefeng Yan \and Mingqiang Wei \at
              School of Computer Science and Technology, Nanjing University of Aeronautics and Astronautics, Nanjing, China \\
         Fu Lee Wang \at School of Science and Technology, Hong Kong Metropolitan University, Hong Kong, China \\
         Haoran Xie \at Department of Computing and Decision Sciences, Lingnan University, Hong Kong, China \\
         Wenhan Yang \at School of Electrical and Electronic Engineering, Nanyang Technological University, Singapore \\
         Jing Qin \at School of Nursing, The Hong Kong Polytechnic University, Hong Kong, China \\
%
}
\date{Received: date / Accepted: date}

\maketitle

\begin{abstract}
\sloppy{  
While the wisdom of training an image dehazing model on synthetic hazy data can alleviate the difficulty of collecting real-world hazy/clean image pairs, it brings the well-known domain shift problem.
From a different yet new perspective, this paper explores contrastive learning with an adversarial training effort to leverage unpaired real-world hazy and clean images, thus bridging the gap between synthetic and real-world haze is avoided.
We propose an effective unsupervised contrastive learning paradigm for image dehazing, dubbed UCL-Dehaze. 
Unpaired real-world clean and hazy images are easily captured, and will serve as the important positive and negative samples respectively when training our UCL-Dehaze network. To train the network more effectively, we formulate a new self-contrastive perceptual loss function, which encourages the restored images to approach the positive samples and keep away from the negative samples in the embedding space.
Besides the overall network architecture of UCL-Dehaze, adversarial training is utilized to align the distributions between the positive samples and the dehazed images.
Compared with recent image dehazing works, UCL-Dehaze does not require paired data during training and utilizes unpaired positive/negative data to better enhance the dehazing performance.
We conduct comprehensive experiments to evaluate our UCL-Dehaze and demonstrate its superiority over the state-of-the-arts, even only 1,800 unpaired real-world images are used to train our network. \textit{Source code has been available at \textcolor{magenta}{ \href{https://github.com/yz-wang/UCL-Dehaze}{https://github.com/yz-wang/UCL-Dehaze}}}.}
\end{abstract}

\section{Introduction}
Images captured by outdoor vision systems often suffer from noticeable degradation of visibility and contrast due to the existence of haze.
Such hazy images inevitably deteriorate the performance of various high-level vision tasks, e.g., traffic monitoring, object detection, and outdoor surveillance \citep{choi2017sharpness,chen2018domain,nasir2019fog}. 

\begin{figure*}[htbp] \centering
	\includegraphics[width=1.0\linewidth]{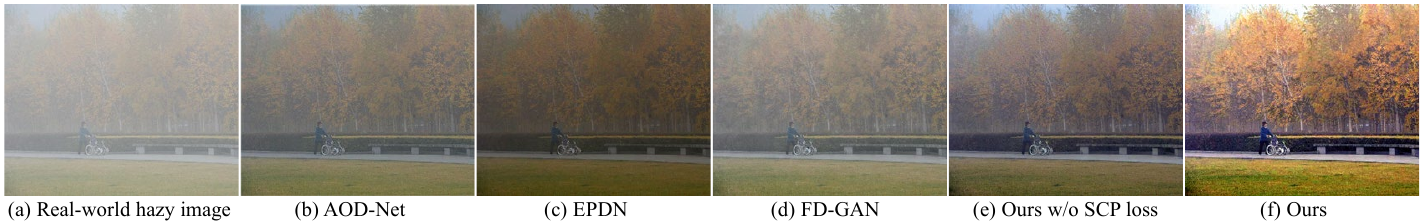}
	\caption{Our UCL-Dehaze avoids bridging the gap between synthetic and real-world haze. It exploits contrastive learning with an adversarial training
effort to leverage unpaired real-world hazy (negatives) and clean (positives) images. Thus, when inputting an unpredictable real-world haze image, UCL-Dehaze often generates a haze-free yet perceptually more pleasing result (f), compared to (b) AOD-Net \citep{li2017aod}, (c) EPDN \citep{qu2019enhanced}, (d) FD-GAN \citep{dong2020fd}, and (e) UCL-Dehaze w/o the self-contrastive perceptual (SCP) loss}
	\label{fig:fig1}
\end{figure*}

Image dehazing aims to recover sharp images from their hazy counterparts, which is a typical ill-posed problem. 
To make this problem well-posed, conventional efforts usually exploit hand-crafted priors with empirical observations, such as the dark channel prior (DCP) \citep{he2011single}, color attenuation prior (CAP) \citep{zhu2015fast}, and non-local color prior (NCP) \citep{berman2016non}, etc. 
Although these methods improve the overall visibility of hazy images, making use of a particular prior assumption to dehaze arbitrary real-world images may not always produce satisfactory results. In addition, even professional users often have to carefully tweak various parameters in the formulas for dehazing different input images. 

To overcome the aforementioned problems, numerous learning-based methods have been proposed \citep{ren2016single,li2017aod,pang2018visual,DBLP:journals/tmm/LiGGHFC20,qin2020ffa,song2020simultaneous,li2021deep}. They commonly employ CNNs or GANs to restore clean images from the corresponding hazy inputs under full supervision or even semi-supervision.
Theoretically, if fed with enough paired data, these (semi-)supervised paradigms may generate very promising dehazing results. 
However, from a practical view, such paired data in the real world are difficult or even impossible to obtain. This explains why existing approaches resort to synthetic hazy data for training. 
But the gap between synthetic and real-world hazy images inevitably degrades their dehazing abilities to deal with real-world scenarios. 
Additionally, most of these learning-based methods only exploit clean images as positive samples to guide the network's training, while ignoring the fact that the unexploited information in hazy images is valuable as negative samples. That is, these negative samples also provide beneficial supervision information to improve the performance of cutting-edge dehazing models.
As exhibited in Fig. \ref{fig:fig1}, compared with the state-of-the-art dehazing approaches and our partial scheme that all only adopt the positive samples, the proposed UCL-Dehaze with the additional negative samples produces a much clearer and perceptually more pleasing dehazing result.


We propose a novel unsupervised contrastive learning paradigm, which casts real-world image dehazing as an image-to-image translation problem (termed UCL-Dehaze). 
UCL-Dehaze builds itself on the contrastive learning framework and benefits from adversarial training efforts.
To effectively train the network in an unsupervised manner, in addition to the patch-wise contrastive loss \citep{park2020contrastive}, we formulate a new pixel-wise contrastive loss, i.e., the self-contrastive perceptual (SCP) loss to encourage the restored images and the clean images (positive samples) to pull together in the representation space while pushing them away from the hazy ones (negative samples).
Both quantitative and qualitative results prove that our UCL-Dehaze performs favorably against the state-of-the-art dehazing approaches, even only 1,800 unpaired real-world training images are used.

In summary, the contributions of this work are three-fold:

\begin{itemize}
\item[$\bullet$]	
We propose an unsupervised image dehazing network via contrastive learning and adversarial training (call UCL-Dehaze). UCL-Dehaze leverages real-world hazy images as negative samples to provide additional supervision information for the network's training. It can effectively address unpredictable real-world hazy scenarios.
\item[$\bullet$] 
We formulate an effective pixel-wise self-contrastive perceptual (SCP) loss to train UCL-Dehaze. Specifically, we employ SCP to learn a representation that pulls the restored images and clean images (positives) together while pushing them away from the hazy ones (negatives). Moreover, SCP can be regarded as a universal module to enhance the performance of any other unsupervised dehazing approaches.
\item[$\bullet$] 
Our UCL-Dehaze is compared with 18 representative state-of-the-art dehazing approaches via comprehensive experiments. The results are evaluated in terms of full-, reduced- and no-referenced image quality assessment, visual quality, and human subjective surveys. Consistently and substantially, UCL-Dehaze performs favorably against SOTAs.
\end{itemize}
\section{Related Work}
Image dehazing can be roughly divided into two categories: prior-based and learning-based approaches. In this section, we briefly introduce these two categories, followed by the introduction of contrastive learning.
\subsection{Single Image Dehazing}
\textbf{Prior-based:} Conventional dehazing methods commonly explore hand-crafted priors to restore haze-free images based on the ill-posed atmospheric scattering model \citep{narasimhan2000chromatic}. \cite{tan2008visibility} develop a dehazing approach via compensating the local contrast of the hazy images. \cite{he2011single} propose the well-known dark channel prior (DCP) for single image dehazing, which achieves impressive dehazing results. Recently, \cite{berman2016non} observe that the colors of a haze-free image can be well approximated by several hundred distinct colors, and then exploit the non-local color prior (NCP)-based dehazing method. Although these methods have achieved promising results, their performances are limited by the accuracy of the hand-crafted priors adopted in the various real-world scenarios.

\textbf{Learning-based:} With the advances in deep learning, various learning-based models have been exploited for image dehazing \citep{ren2016single,li2017aod,yin2019color,ren2018gated,liu2019griddehazenet,DBLP:conf/cvpr/DongPXHZ0020,chen2021psd}. Early efforts focus on employing CNNs to estimate the transmission map and global atmospheric light in the atmospheric scattering model and then generate haze-free images. For instance, the MSCNN proposed by \cite{ren2016single} is one of the early approaches that adopts CNNs for single image dehazing, where the model is trained to estimate the transmission map and then restore the clean result. Recently, some other learning-based approaches have tried to directly produce haze-free images in an end-to-end manner. \cite{li2017aod} develop a novel all-in-one dehazing network termed AOD-Net, which is the first model to directly learn the hazy-to-clean image translation. Since then, numerous end-to-end dehazing methods have sprung up. \cite{qin2020ffa} exploit a novel feature fusion network to directly restore haze-free images from the hazy inputs, which achieves remarkable performance on several benchmark datasets. Although we have witnessed promising dehazing results on synthetic data- sets, these learning-based efforts trained on synthetic images cannot perform well under real-world scenarios due to the obvious domain gap. 

Recently, this issue has been picked up by several semi-supervised-based studies \citep{li2020semi,an2021semi}. These approaches are explored to train their models on both synthetic data and real-world images. Although they alleviate the problem of domain shifts to some extent, their dehazing abilities still depend on the quality and quantity of synthetic data.


Motivated by the success of CycleGAN in unpaired image-to-image translation \citep{zhu2017unpaired}, a handful of approaches attempt to exploit unsupervised frameworks for image dehazing in order to solve this domain shift issue (e.g., Cycle-Dehaze \citep{engin2018cycle}, Dehaze-GLCGAN \citep{anvari2020dehaze}). Since such models can leverage unpaired images for network training, they commonly generalize well on real-world scenes. 
However, most learning-based efforts only employ clean images as positive samples to guide network training, while ignoring the fact that the negative samples can also provide additional beneficial information for the network's training, thus limiting the dehazing performance of the model.

\subsection{Contrastive Learning}
Contrastive learning is a kind of self-supervised learning framework, and it is widely used in the representation learning field \citep{sermanet2018time,oord2018representation,he2020momentum,chen2020simple,henaff2020data}. These approaches aim to learn an embedding that brings the associated features close to each other, while the irrelevant samples are pushed away. Existing efforts mainly focus on applying the contrastive learning paradigm on high-level vision tasks, since the data augmentation method is very suitable for modeling the contrast between positive and negative samples. Recently, inspired by the success of contrastive learning in high-level vision tasks, several 
studies have attempted to apply contrastive learning to low-level vision tasks. For instance, \citet{han2021underwater} propose a contrastive learning framework for underwater image restoration. \citet{wu2021contrastive} develop a contrastive regularization term to leverage the information of both hazy and clean images for image dehazing. These approaches demonstrate the great potential of the contrastive learning paradigm in improving the performance of low-level vision tasks. 


Different from the previous works, we do not plan to bridge the gap between synthetic and real-world haze. Thus, we explore unsupervised contrastive learning from an adversarial training perspective to leverage unpaired real-world hazy and clean images. Our proposed network does not require paired data during training. By training the network both pixel-wisely and patch-wisely in an unsupervised yet adversarial manner, we can better utilize unpaired positive/ negative data to enhance its dehazing performance.

\section{UCL-Dehaze}
In this section, we first describe the overview of UCL-Dehaze, and then detail its architecture. After that, we introduce how the patch-wise and pixel-wise contrastive learning paradigm with an adversarial training effort works in UCL-Dehaze.
\subsection{Overview}
In our design, we cast image dehazing as an image-to-image translation task, and simplify it by seeking a mapping function from the hazy images to the corresponding haze-free images. 
However, considering that the gap between synthetic and real-world data may cause a significant dehazing performance drop in real-world scenarios, we focus on developing an unsupervised learning framework to train our network via unpaired real-world images rather than paired synthetic images with `fake' haze. 
In this way, abundant practical real-world images can contribute to the network's training, thus boosting the generalization ability of our network on real-world hazy images. 

\begin{figure*}[!ht]
	\centering
	\includegraphics[width=0.9\linewidth]{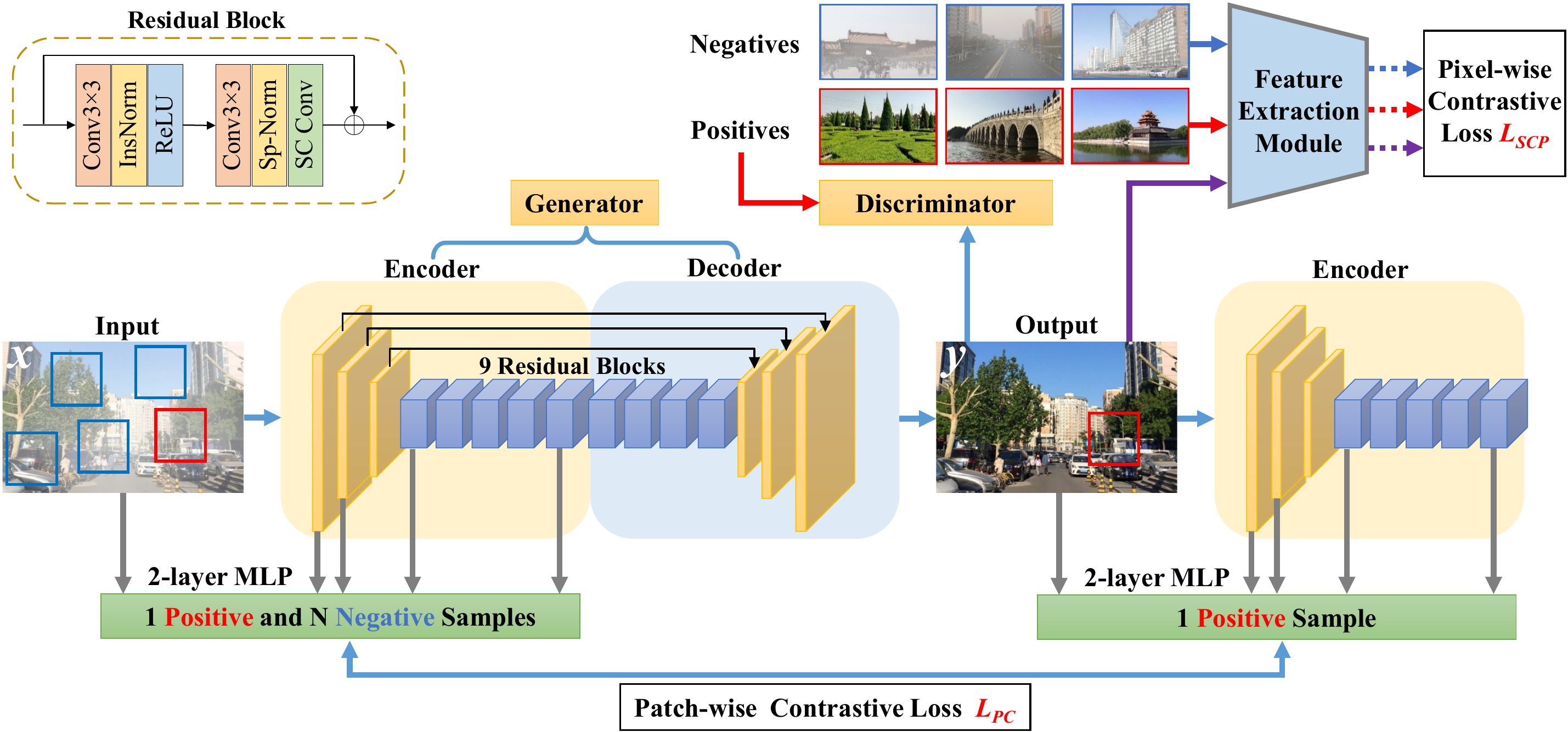}
    \caption{Overview of UCL-Dehaze. UCL-Dehaze aims to learn a mapping function $F$, which can map the input hazy image $x$ to the corresponding haze-free image $y$ in an unsupervised manner. It leverages both patch-wise and pixel-wise contrastive losses for the network's training, thus enhancing the dehazing ability of UCL-Dehaze, especially in real-world scenarios. $L_{SCP}$ represents self-contrastive perceptual loss, Sp-Norm refers to spectral Normalization, and SC Conv refers to self-calibrated convolutions}
	\label{overview}
\end{figure*}

Towards real-world image dehazing, we propose an unsupervised contrastive learning paradigm, called UCL-Dehaze. As exhibited in Fig. \ref{overview}, the overall architecture of UCL-Dehaze is a UNet-like generator \citep{ronneberger2015u} with nine residual blocks \citep{he2016deep}. Given an input real-world hazy image $x$, we aim to employ the generator $G$ to map $x$ to the haze-free image $y$ in an unsupervised training manner. To this end, we leverage contrastive learning with an adversarial training strategy for network training. Specifically, we first feed $x$ to $G$ to produce the preliminary dehazing result $y$. Then, we employ the discriminator $D$ to judge whether $y$ is a real clean image or a fake image produced by $G$, which can further improve the quality of $y$. Finally, we leverage both patch-wise and pixel-wise contrastive losses to train the network in an unsupervised manner, so as to handle unpredictable real-world hazy scenes.

\subsection{Network Architecture}
We employ a UNet-based network with nine residual blocks as the generator module. 
Although using a more complex network structure would improve the dehazing performance of the model, we choose to adopt a simple ResNet-based generator to achieve a better parameter-performance trade-off. 
As known, the training process of GANs is very unstable, and problems such as mode collapse and convergence difficulties often occur. We employ the spectral normalization strategy \citep{miyato2018spectral} in the design of residual blocks, which enhances the stability of training. In addition, to further boost the dehazing performance of UCL-Dehaze, an up-to-date multi-scale feature extraction module (self-calibrated convolutions, i.e., SC Conv) developed by \cite{liu2020improving} is introduced into our network.

\textbf{Generator.} As demonstrated in Fig. \ref{overview}, given an input hazy image $x$, the generator $G$ can map $x$ to the haze-free image $y$ in an end-to-end manner. To achieve this goal, $G$ is supposed to preserve both image structures and details when removing the haze. Motivated by previous studies, we exploit an encoder-decoder network with nine residual blocks as the generator, and introduce the skip connection mechanism to avoid gradient vanishing. 

Given a hazy image, we first employ a 4× down-sampling operation to encode the input hazy image into a low-resolution feature map. Then, nine residual blocks are adopted to extract more complex and deeper features in the low-resolution space and remove the haze simultaneously. After that, we employ the corresponding 4× up-sampling operation and a 7×7 convolutional layer to output the final restored image. Moreover, as mentioned above, we introduce the SC Conv module in the generator design to further enhance the dehazing ability of UCL-Dehaze.

\textbf{Self-calibrated Convolutions.} The self-calibrated convolution module is an improved CNN architecture that can capture long-distance spatial and inter-channel dependencies around each spatial location. Therefore, it can expand the receptive field of each convolutional layer and help CNNs to produce richer features. Given this, we employ the self-calibrated convolution module as a multi-scale feature extraction module to improve the dehazing performance of UCL-Dehaze. 

\begin{figure}[!ht]
	\centering
	\includegraphics[width=1.0\linewidth]{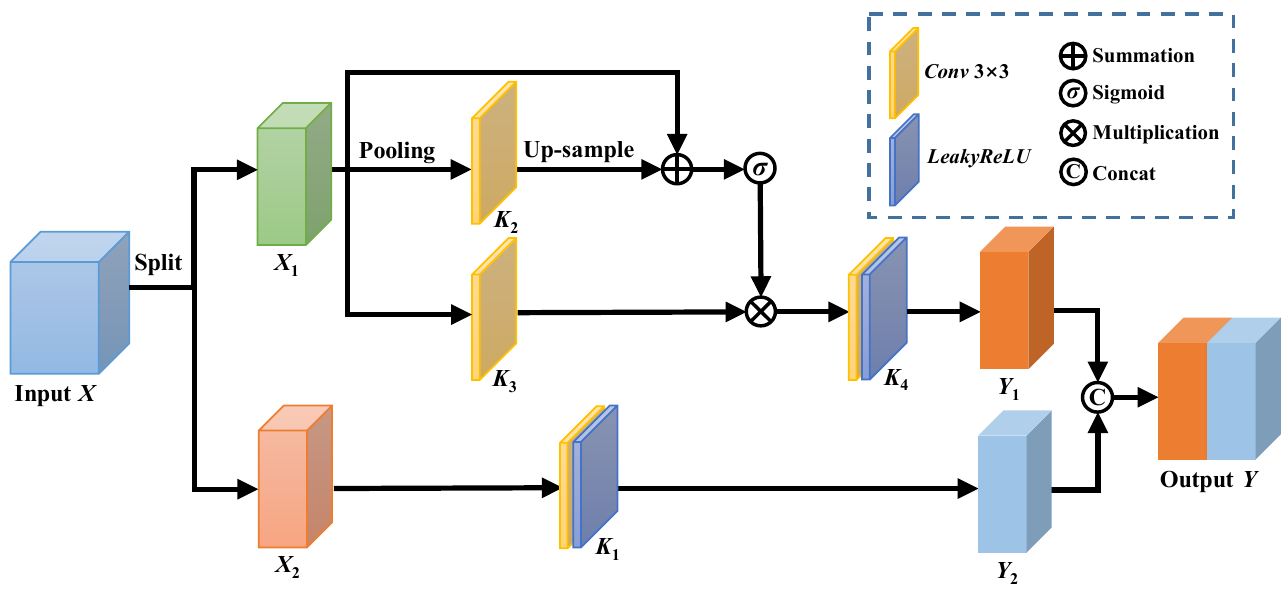}
    \caption{Architecture of Self-calibrated Convolutions}
	\label{fig3}
\end{figure}

As demonstrated in Fig. \ref{fig3}, given an input feature map $X$, we first split it into two feature maps $X_1$ and $X_2$. Then, the self-calibrated convolution module leverages four different convolution operations (i.e., $K_1$, $K_2$, $K_3$ and $K_4$) to extract and fuse multi-scale features from $X_1$ and $X_2$, thus enriching their feature representations. After that, we can obtain the output features $Y_1$ and $Y_2$ from the two branches of self-calibrated convolutions. Finally, we concatenate $Y_1$ and $Y_2$ to produce the final output $Y$. In our design, we add the self-calibrated convolution module after the ReLU layer of each convolution operation to expand the receptive field of the convolutional layer and extract multi-scale features, thereby boosting the performance of the generator.

\textbf{Discriminator.} For adversarial training, we employ the well-known PatchGAN \citep{isola2017image} as the discriminator, which can reduce the network's parameters and perform faster than other conventional discriminators. 
The function of the discriminator is to judge whether a given image is a real clean image or a fake image produced by the generator, thus guiding the generator to produce more realistic images. Least-Square GAN (LSGAN) loss \citep{mao2017least} has been proved to be more effective than the vanilla GAN loss, as it can ensure that the training process to be more stable. We adopt the LSGAN loss to train our network. The definition of adversarial loss can be expressed as:
\begin{equation}\label{eq:advg}
\begin{aligned}
L_{adv}(G)=E_{G(x) \sim P_{fake}}[(D(G(x))-1)^2],
\end{aligned}
\end{equation}
\begin{equation}\label{eq:advd}
\begin{aligned}
L_{adv}(D)=E_{y \sim P_{real}}[(D(y)-1)^2]\\
+E_{G(x) \sim P_{fake}}[(D(G(x)))^2],
\end{aligned}
\end{equation}
where $y$ refers to the real-world clean images and $G(x)$ represents the restored haze-free images.

\subsection{Patch-wise Contrastive Learning}
Contrastive learning aims to learn an embedding to push the positive samples close to each other and push apart the embedding between negative samples. 
We leverage contrastive learning to train our network in an unsupervised manner. The first thing we need to consider is how to construct the positive and negative samples. 
Inspired by CUT \citep{park2020contrastive} and CWR \citep{han2021underwater}, we randomly choose $N+1$ patches from the input image $x$ and one corresponding patch from the restored image $y$ (see Fig. \ref{overview}). 
We denote the two corresponding patches as the positive sample, while the other N patches in $x$ are the negative samples. Then, we employ a noisy contrastive estimation module to maximize the mutual information between positive samples (corresponding patches in $x$ and $y$). Specifically, we first map the anchor (the patch in $y$), the positive and the $N$ negatives to $K$-dimensional vectors and denote them as $v$, $v^+$, and $v^-$, respectively. After that, we establish an ($N+1$)-way classification problem and compute the probability of $v^+$ being selected over $v^-$, which is formulated as a cross-entropy loss:
\begin{equation}\label{patchNCE}
\begin{array}{l}
\ell\left({v}, {v}^{+}, {v}^{-}\right)= \\
-\log \left(\frac{\exp \left({sim}\left(v, {v}^{+}\right) / \tau\right)}{\exp \left({sim}\left(v, {v}^{+}\right) / \tau\right)+\sum_{n=1}^{N} \exp \left({sim}\left(v, {v}_{n}^{-}\right) / \tau\right)}\right),
\end{array}
\end{equation}
where $sim(u,v)$ refers to the cosine similarity between $u$ and $v$. $\tau$ is a temperature parameter to adjust the distance between the anchor and other samples and is set to 0.07 as the default. ${v}_{n}^{-}$ denotes the n-th negative sample.

We adopt the encoder module $G_{enc}$ and a two-layer MLP network $H$ to extract features from the input image $x$ (see Fig. \ref{overview}), and embed them to a stack of features $\left\{z_{l}\right\}_{L}=\left\{H^{l}\left(G_{\text {enc }}^{l}(x)\right)\right\}_{L}$, where $L$ represents the number of layers we choose from the $G_{enc}$, and $l$ is the specific $l$-th selected layers. These stack of features actually represent different patches from the image, and we denote the spatial locations in each selected layer as $s \in\left\{1, \ldots, S_{l}\right\}$, where $S_{l}$ refers to the number of spatial locations in each layer. We select an anchor each time and denote its feature as $\hat{z}_{l}^{s} \in \mathbb{R}^{C_{l}}$, where $C_{l}$ is the number of channels in each layer. Additionally, we refer to the corresponding feature (i.e., positive) as ${z}_{l}^{s} \in \mathbb{R}^{C_{l}}$ and the other features (i.e., negatives) as ${z}_{l}^{S \backslash s} \in \mathbb{R}^{\left(S_{l}-1\right) \times C_{l}}$. 

Our goal is to match the corresponding patches (positives) of input and output images while pushing the other patches (negatives) away from the anchor. Therefore, the patch-wise, multi-layer contrastive loss for mapping $X \rightarrow Y$(i.e., hazy images $\rightarrow$ clean images) can be formulated as: 
\begin{equation}\label{patchNCE}
\begin{array}{l}
{L}_{PC}(G, H, X)=\mathbb{E}_{{x} \sim X} \sum_{l=1}^{L} \sum_{s=1}^{S_{l}} \ell\left(\hat{z}_{l}^{s}, {z}_{l}^{s}, {z}_{l}^{S \backslash s}\right).
\end{array}
\end{equation}

\subsection{Pixel-wise Contrastive Learning}
To better restore the clean images from their hazy counterparts, we adopt the real-world clean images and hazy images as the positive and negative samples to reconstruct the sharp images from the corresponding hazy images. Note that all these positive/negative samples are randomly chosen from the real-world images and are unpaired from each other. We aim to encourage the restored images to be close to the positive samples while keeping away from the negative ones in the embedding space. 

Motivated by AECR-Net \citep{wu2021contrastive}, we develop a novel pixel-wise self-contrastive perceptual loss to achieve this goal. In our method, we denote the group of a real-world clean image $R_c$ and a preliminary restored image $G(x)$ as the positive pair. Similarly, the negative pair is generated by the group of a real-world hazy image $R_h$ and a restored image $G(x)$. In addition to constructing the positive and negative pairs, we need to find a latent feature space of these pairs for contrast. Here, we employ a pre-trained VGG-16 network to extract the feature maps of different samples. Therefore, the pixel-wise contrastive loss can be expressed as:
\begin{equation}\label{patchNCE}
\begin{array}{l}
L_{SCP} =\lambda \sum_{i=1}^{n} \omega_{i} \cdot \frac{\left\|\psi_i\left(R_c\right)-\psi_i(G(x))\right\|_{1}}{\left\|\psi_i\left(Rh\right)-\psi_i(G(x))\right\|_{1}},
\end{array}
\end{equation}
where $\psi_i(.),i=1,2, \cdots n$, refer to extracting the $i$-th hidden features from the VGG-16 network pre-trained on ImageNet. Here we choose the 2-nd, 3-rd and 5-th max-pooling layers. $\omega_{i}$ are weight coefficients, and we set $\omega_{1}$ = 0.4, $\omega_{2}$ = 0.6, and $\omega_{3}$ = 1. Besides these two contrastive loss functions, we also adopt the identity loss to keep the structure identical after dehazing, which is shown as the following formula: 
\begin{equation}\label{patchNCE}
\begin{array}{l}
L_{ide}=E_{y \sim P_{data(Y)}}\left[\left\|\left(G({y})-y\right)\right\|_{1}\right].
\end{array}
\end{equation}

Such an identity loss can encourage the output image to have the same color composition and structure as the input image, thus enhancing the quality of the generated image. The total loss function can be formulated as:
\begin{equation}\label{patchNCE}
\begin{array}{l}
L_{{Total }}=\lambda_{1} L_{adv}(G)+\lambda_{2} L_{PC}+\lambda_{3} L_{SCP}+\lambda_{4} L_{{ide}},
\end{array}
\end{equation}
where $\lambda_{i}, i=1,2, \cdots 4$, are hyperparameters, and we set $\lambda_{1}=1$, $\lambda_{2}=1$, $\lambda_{3}=0.0002$ and $\lambda_{4}=5$ in our experiments.

\section{Experiments}
In this section, comprehensive experiments are conducted to evaluate the dehazing performance of UCL-Dehaze and other methods. All the experiments are implemented by PyTorch 1.7 on a system with an Intel(R) Core(TM) i9-10920X CPU and an NVIDIA GeForce RTX 3090 GPU.

\subsection{Implementation Details}

\textbf{Dataset.} Since UCL-Dehaze is trained in an unsupervised manner, real-world hazy images can contribute to the network's training for real-world scenarios. 
We randomly choose the real-world hazy and clean images from the most accessible publicly dataset RESIDE \citep{li2018benchmarking} as our training set. 
RESIDE is a widely used benchmark dataset for image dehazing, which consists of six subsets, i.e., OTS (Outdoor Training Set), ITS (Indoor Training Set), SOTS (Synthetic Object Testing Set), RTTS (Real-world Task-driven Testing Set), HSTS (Hybrid Subjective Testing Set), and URHI (Unannotated Real Hazy Images). In our experiments, the training set is composed of 1,800 real-world hazy images chosen from RTTS and URHI. For haze-free images, we randomly choose 1,800 clean images from ITS. Note that all the hazy and clean images in our training set are real-world images and are unpaired from each other. Both the SOTS and HSTS are adopted as the testing set. Although increasing the number of images in the training set can further improve the dehazing performance of UCL-Dehaze, even in this case, our approach surpasses various state-of-the-art dehazing approaches trained on the entire ITS (including 100,000 indoor hazy/clean images). 

\textbf{Training Details.} UCL-Dehaze is trained using the Adam optimizer \citep{kingma2014adam} with a batch size of 1, where the momentum parameters $\beta_{1}$ and $\beta_{2}$ are set to 0.5 and 0.999, respectively. The initial learning rate $l$ for both generator and discriminator is set to $2 \times 10^{-{4}}$. We empirically set the total number of epochs to 100 and adopt a linear decay strategy to adjust $l$ after 50 epochs. Additionally, the patch-wise contrastive loss $L_{PC}$ is computed by the features from five layers of encoder $G_{enc}$, i.e., the input RGB image, the 1-st and 2-nd down-sampling convolutional layers, as well as the 1-st and 5-th residual blocks. For each layer's features, we randomly sample 256 locations and apply a 2-layer MLP module to produce the final 256-dim features.

Moreover, we are surprised to find that if we employ $L_{PC}$ for mapping both $X \rightarrow Y$ and $Y \rightarrow Y$ (i.e., clean images $\rightarrow$ clean images), the restored images will be much clearer and more realistic. Therefore, we adopt a dual-direction $L_{PC}$ to train our UCL-Dehaze, and the two hyperparameters $\lambda_{2}$ are both set to 1. 

\begin{table*}[htbp]
	\centering
	\caption{Quantitative PSNR and SSIM values of the proposed UCL-Dehaze and 18 state-of-the-art dehazing approaches on synthetic datasets. Our UCL-Dehaze achieves the best performance}
	\label{Tab1}
	\begin{threeparttable}
		\footnotesize
		\centering
		\setlength{\tabcolsep}{0.9mm}{
			\begin{tabular}{lllllll}
				\toprule
				\multirow{2}{*}{Method}&
				\multirow{2}{*}{Publication}&
				\multirow{2}{*}{Type}&
				\multicolumn{2}{c}{SOTS outdoor}&
				\multicolumn{2}{c}{HSTS}\cr
				 \cmidrule(lr){4-5} \cmidrule(lr){6-7}&
				 & & PSNR$\uparrow$ & SSIM$\uparrow$ & PSNR$\uparrow$ & SSIM$\uparrow$ \cr
				\midrule
				DCP \citep{he2011single}  & TPAMI'11 & Prior & 18.38 & 0.819 & 17.01 & 0.803 \cr
				BCCR \citep{meng2013efficient}  & ICCV'13  & Prior & 15.71 & 0.769 & 15.21 & 0.747 \cr
				NCP \citep{berman2016non}  & CVPR'16 & Prior & 18.07 & 0.802 & 17.62 & 0.798 \cr
				AOD-Net \citep{li2017aod}  & ICCV'17 & Supervised & 20.08 & 0.861 & 19.68 & 0.835 \cr
				GFN \citep{ren2018gated} & CVPR'18 & Supervised & 21.49 & 0.838 & 22.94 & 0.894 \cr
				EPDN \citep{qu2019enhanced}  & CVPR'19 & Supervised & 22.57 & 0.863 & 20.37 & 0.877 \cr
				GCANet \citep{chen2019gated} & WACV'19 & Supervised & 21.66 & 0.867 & 21.37 & 0.874 \cr
				MSCNN-HE \citep{ren2020single} & IJCV'20  & Supervised & 22.72 & 0.871 & 21.23 & 0.851 \cr
				FD-GAN \citep{dong2020fd}  & AAAI'20  & Supervised & 23.76 & 0.926 & 23.28 & 0.914 \cr
				GFN-IJCV \citep{zhang2020gated}  & IJCV'20  & Supervised & 24.21 & 0.849 & 23.17 & 0.829 \cr
				Interleaved CSF \citep{wu2020learning}  & TIP'20  & Supervised & 24.17 & 0.923 & 22.94 & 0.907 \cr
				Semi-dehazing \citep{li2020semi}  & TIP'20  & Semi-supervised & 24.79 & 0.892 & 24.36 & 0.889 \cr
				CycleGAN \citep{zhu2017unpaired} & ICCV'17 & Unsupervised & 17.32 & 0.706 & 16.05 & 0.703 \cr
				Cycle-Dehaze \citep{engin2018cycle}  & CVPRW'18 & Unsupervised & 18.60 & 0.797 & 17.96 & 0.777 \cr
				Deep DCP \citep{golts2020unsupervised}  & TIP'20 & Unsupervised & 20.99 & 0.893 & 21.21 & 0.871 \cr
				LIGHT-Net \citep{dudhane2020end} & TETCI'20 & Unsupervised & 23.11 & 0.917 & 22.27 & 0.906 \cr
				YOLY \citep{li2021you}  & IJCV'21  & Unsupervised & 20.39 & 0.889 & 21.02 & 0.905 \cr
				PSD \citep{chen2021psd}  & CVPR'21 & Unsupervised & 20.49 & 0.844 & 19.37 & 0.824 \cr
				UCL-Dehaze & Ours & Unsupervised & \textbf{25.21} & \textbf{0.927} & \textbf{26.87} & \textbf{0.933} \cr
				\bottomrule
			\end{tabular}
		}
	\end{threeparttable}
\end{table*}

\textbf{Evaluation Settings.} UCL-Dehaze is compared quantitatively and qualitatively with various dehazing approaches. They can be classified into three categories: 1) prior-based DCP \citep{he2011single}, BCCR \citep{meng2013efficient} and NCP \citep{berman2016non}) supervised-based AOD-Net \citep{li2017aod}, GFN \citep{ren2018gated}, EPDN \citep{qu2019enhanced}, GCANet \citep{chen2019gated}, MSCNN-HE \citep{ren2020single}, FD-GAN \citep{dong2020fd}, GFN-IJCV \citep{zhang2020gated}, and Interleaved CSF \citep{wu2020learning}; and 3) unsupervised-based CycleGAN \citep{zhu2017unpaired}, Cycle-Dehaze \citep{engin2018cycle}, Deep DCP \citep{golts2020unsupervised}, LIGHT-Net \citep{dudhane2020end}, YOLY \citep{li2021you}, and PSD \citep{chen2021psd}. 
Besides, we also compared UCL-Dehaze with a recent semi-supervised image dehazing framework called Semi-dehazing \citep{li2020semi}.
We employ the average Peak Signal to Noise Ratio (PSNR) and Structural Similarity index (SSIM) for quantitative evaluation of the recovered images, which are the most widely used image full-reference evaluation indexes.
We also leverage the well-known CIED- E2000 \citep{sharma2005ciede2000} to measure the color difference between the restored image and its haze-free counterpart. Furthermore, to assess the quality of the dehazed images more comprehensively, four reduced-reference indicators are employed to evaluate the contrast (Contrast gain \citep{DBLP:journals/ivc/EconomopoulosAM10}), visibility ($e, \bar{r}$) \citep{hautiere2008blind} and saturation ($\sigma$) \citep{hautiere2008blind} of the restored images.

1). Contrast gain refers to the mean contrast difference between the dehazing image and its hazy counterpart, which is formulated as: 
\begin{equation}
C_{\text {gain }}=\bar{C}_{R}-\bar{C}_{H},
\end{equation}
where $\bar{C}_{R}$ and $\bar{C}_{H}$ are the mean contrast of the restored image and hazy image respectively. Given an image with the size of $N_x \times N_y$, its mean contrast can be expressed by:
\begin{equation}
\bar{C}=\frac{1}{N_{x}N_{y}} \sum_{y=1}^{N_{y}} \sum_{x=1}^{N_{x}} C(x, y),
\end{equation}
where $C$ represents the contrast of the image in a small window and can be calculated by: 
\begin{equation}
C(x, y)=\frac{S(x, y)}{m(x, y)},
\end{equation}
where $S(x, y)=\frac{1}{(2 r+1)^{2}} \sum_{j=-r}^{r} \sum_{i=-r}^{r}(I(x+i, y+j)-m(x, y))^{2}$, $m(x, y)=\frac{1}{(2 r+1)^{2}} \sum_{j=-r}^{r} \sum_{i=-r}^{r} I(x+i, y+j)$. $I(x, y)$ refers to the original hazy image with the size of $N_x \times N_y$. $r$ is the radius of the local region. A larger value of Contrast gain indicates a better result.

2). The indicators ($e, \bar{r}$) evaluate image visibility by measuring the enhanced degree of image edges \citep{hautiere2008blind}. The first indicator $e$ represents the restoration rate of visible edges after image dehazing and can be expressed as:
\begin{equation}
e=\frac{n_{r}-n_{o}}{n_{o}},
\end{equation}
where $n_{r}$ and $n_{o}$ refer to the cardinal numbers of the set of visible edges in the dehazing image $I_r$ and the original image $I_o$. The value of $e$ evaluates the ability of the dehazing algorithm to restore image edges that were not visible in the original hazy image. The second indicator $\bar{r}$ is employed to assess the restoration degree of the image edge and texture information. It takes into account both invisible and visible image edges in $I_o$, which is formulated as:
\begin{equation}
\bar{r}=\exp \left[\frac{1}{n_{r}} \sum_{i \in \wp_{r}} \log r_{i}\right],
\end{equation}
where $r_{i}=\Delta I_{i}^{r} / \Delta I_{i}^{o}$, $\Delta I_{i}^{r}$ and $\Delta I_{i}^{o}$ denote the gradient of the dehazing image and original hazy image, respectively. $\wp_{r}$ refers to the set of visible edges of the restored image. Similar to $e$, a larger $\bar{r}$ means better results.

3). The indicator $\sigma$ is adopted to evaluate the color restoration performance of dehazing methods \citep{hautiere2008blind}. $\sigma$ represents the rate of the saturated pixels (black or white) after image dehazing and can be expressed as:
\begin{equation}
\sigma=\frac{n_{s}}{N_x \times N_y},
\end{equation}
where $n_{s}$ represents the number of pixels that are saturated after applying the image restoration but were not before. A smaller value of $\sigma$ usually indicates a better result.

\begin{table*}[htbp]
	\centering
	\caption{Quantitative comparisons (CIEDE2000/Contrast gain/($e, \bar{r}$)/$\sigma$) with state-of-the-art dehazing algorithms on synthetic datasets. \textcolor{red}{Red} and \textcolor{blue}{blue} colors are used to indicate the $1^{st}$ and $2^{nd}$ ranks, respectively.} 
	\label{Tab22}
	\begin{threeparttable}
		\footnotesize
		\centering
		\setlength{\tabcolsep}{0.9mm}{
			\begin{tabular}{llcccccccccc}
				\toprule
				\multirow{2}{*}{Method}&
				\multirow{2}{*}{Type}&
				\multicolumn{5}{c}{SOTS outdoor}&
				\multicolumn{5}{c}{HSTS}\cr
				 \cmidrule(lr){3-7} \cmidrule(lr){8-12}&
				 & CIDED2000$\downarrow$ & Contrast gain$\uparrow$ & $e$$\uparrow$ & $\bar{r}$$\uparrow$ & $\sigma$$\downarrow$ & CIDED2000$\downarrow$ & Contrast gain$\uparrow$ & $e$$\uparrow$ & $\bar{r}$$\uparrow$ & $\sigma$$\downarrow$ \cr
				\midrule
				DCP  &  Prior & 10.199 & 0.284 & \textcolor{red}{12.285} & 1.439 & 0.0037 & 9.186 & 0.294 & \textcolor{red}{11.108} & 1.473 & 0.0025\cr
				AOD-Net  &  Supervised & 7.287 & 0.230 & 7.755 & 1.673 & \textcolor{blue}{0.0021} & 7.646 & 0.202 & 7.824 & 1.711 & \textcolor{blue}{0.0018} \cr
				GCANet  &  Supervised & 7.314 & 0.210 & 7.274 & 1.568 & 0.0121 & 8.107 & 0.234 & 8.999 & 1.471 & 0.0195 \cr
				FD-GAN  &  Supervised & 6.537 & 0.262 & 8.736 & 1.520 & 0.0204 & 7.122 & 0.297 & 7.247 & \textcolor{blue}{1.892} & 0.0170 \cr
				Semi-dehazing  & Semi-supervised & \textcolor{blue}{4.856} & 0.205 & 8.642 & 1.521 & 0.0124 & \textcolor{blue}{5.312} & 0.191 & 7.270 & 1.628 & 0.0190\cr
				CycleGAN  & Unsupervised & 13.394 & 0.280 & 9.984 & 1.526 & 0.0059 & 15.074 & 0.317 & 7.780 & 1.810 & 0.0023\cr
				Cycle-Dehaze  & Unsupervised & 13.967 & \textcolor{blue}{0.287} & 9.288 & 1.390 & 0.0024 & 13.535 & \textcolor{blue}{0.324} & 8.417 & 1.375 & 0.0034\cr
				PSD  & Unsupervised & 14.292 & 0.281 & 8.721 & \textcolor{red}{3.199} & 0.0185 & 14.820 & 0.272 & 7.866 & \textcolor{red}{3.230} & 0.0021\cr
				UCL-Dehaze & Unsupervised & \textcolor{red}{4.784} & \textcolor{red}{0.289} & \textcolor{blue}{10.685} & \textcolor{blue}{1.685} & \textcolor{red}{0.0008} & \textcolor{red}{4.612} & \textcolor{red}{0.330} & \textcolor{blue}{10.481} & 1.797 & \textcolor{red}{0.0002}\cr
				\bottomrule
			\end{tabular}
		}
	\end{threeparttable}
\end{table*}

\begin{figure*}[htbp] \centering
	\includegraphics[width=1.0\linewidth]{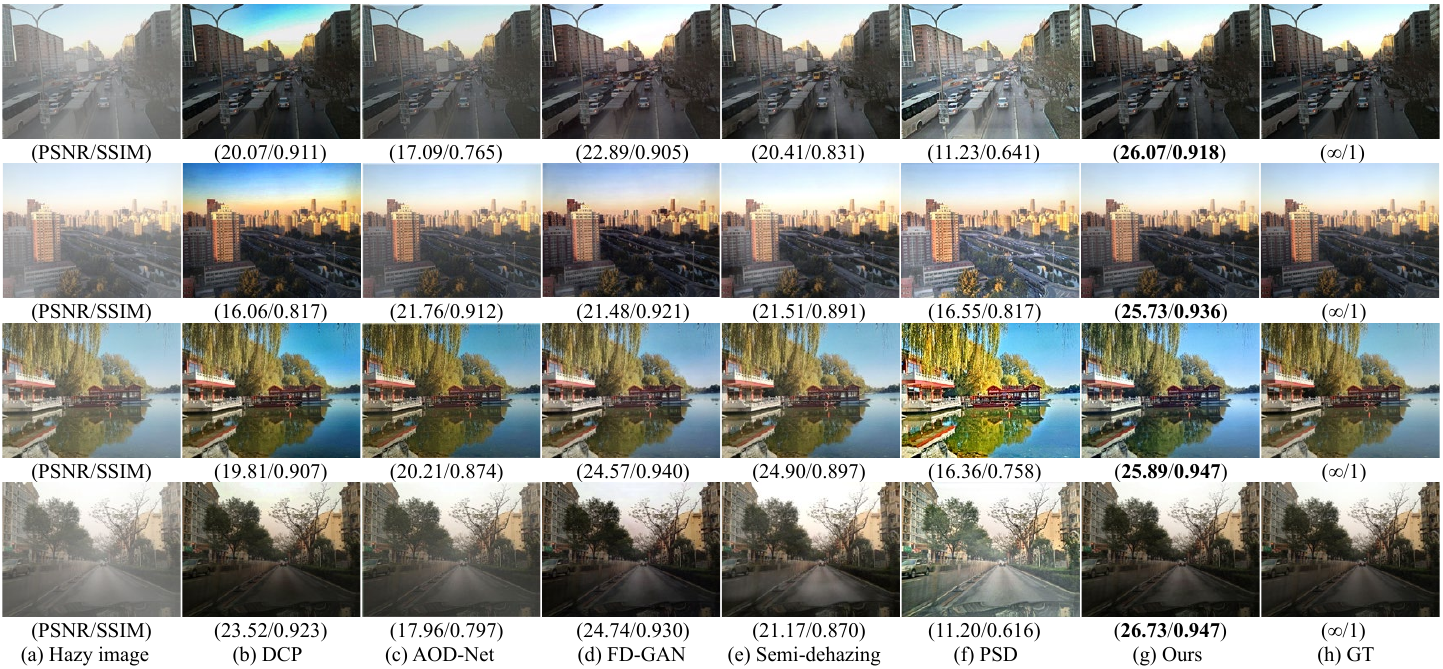}
	\caption{
Image dehazing results on the SOTS outdoor dataset. From (a) to (h): (a) the hazy image, and the dehazing results of (b) DCP \citep{he2011single}, (c) AOD-Net \citep{li2017aod}, (d) FD-GAN \citep{dong2020fd}, (e) Semi-dehazing \citep{li2020semi}, (f) PSD \citep{chen2021psd}, (g) our UCL-Dehaze, respectively, and (h) the ground-truth image. Our UCL-Dehaze can produce much clearer dehazing images with well-preserved details}
	\label{fig:fig4}
\end{figure*}

\begin{figure*}[htbp] \centering
	\includegraphics[width=1.0\linewidth]{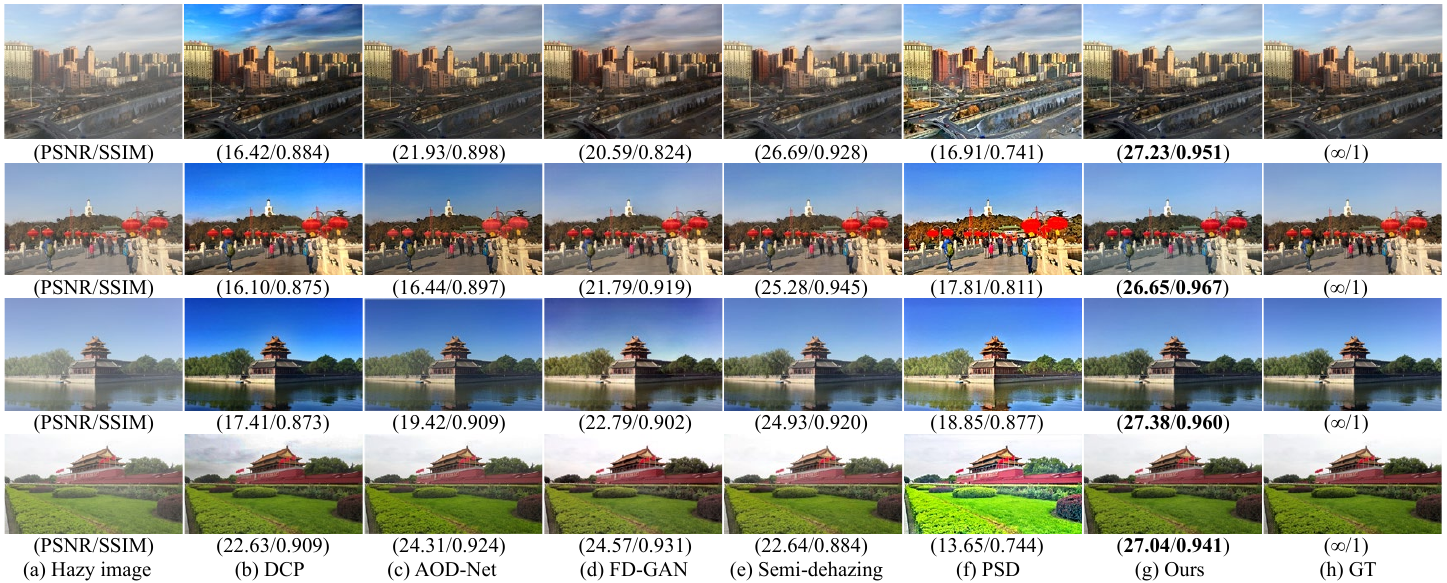}
	\caption{
Image dehazing results on the HSTS dataset. From (a) to (h): (a) the hazy image, and the dehazing results of (b) DCP \citep{he2011single}, (c) AOD-Net \citep{li2017aod}, (d) FD-GAN \citep{dong2020fd}, (e) Semi-dehazing \citep{li2020semi}, (f) PSD \citep{chen2021psd}, (g) our UCL-Dehaze, respectively, and (h) the ground-truth image. UCL-Dehaze can produce much clearer results with perceptually pleasing}
	\label{fig:fig5}
\end{figure*}

\begin{figure*}[htbp] \centering
	\includegraphics[width=1.0\linewidth]{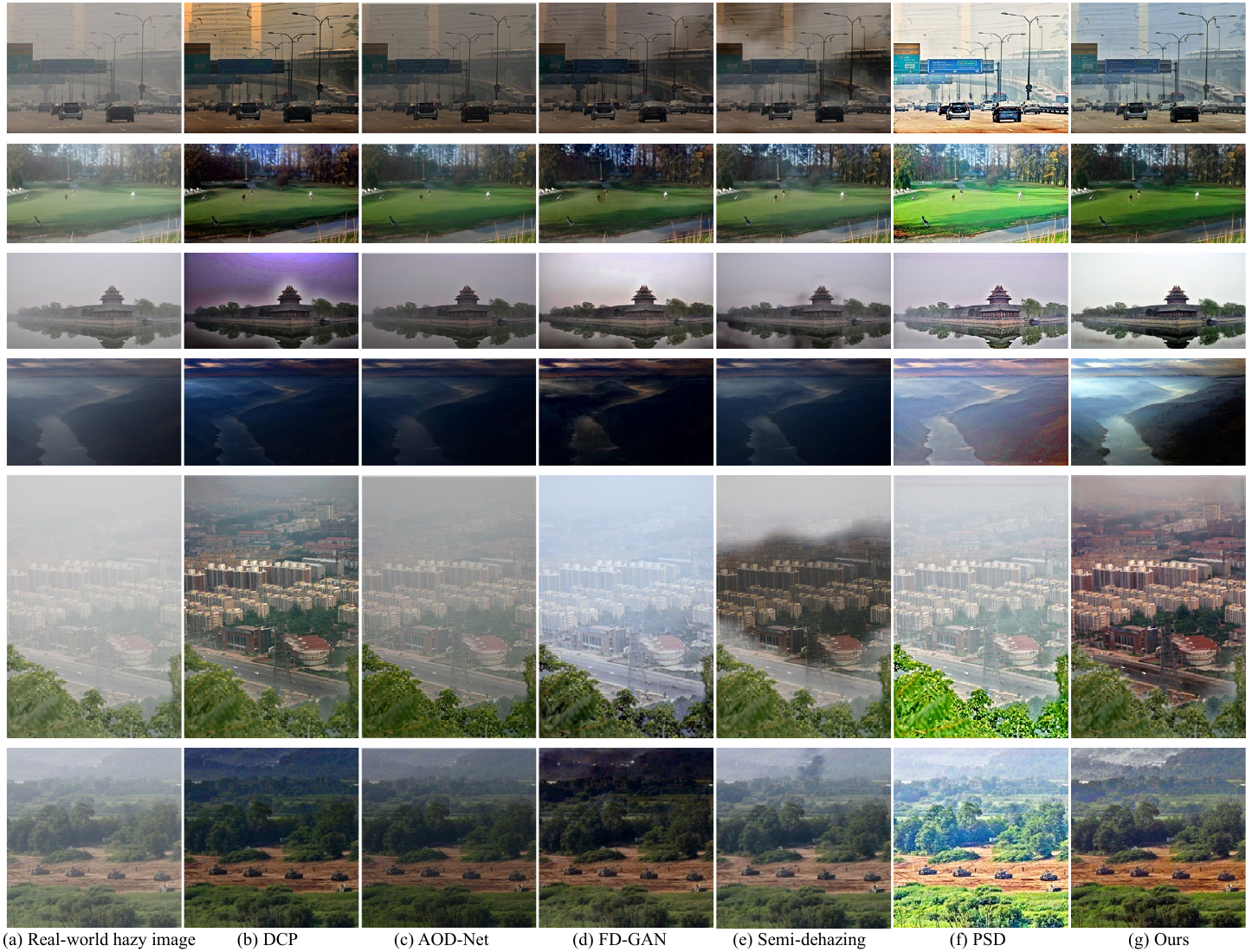}
	\caption{
Image dehazing results on the real-world hazy images. From (a) to (g): (a) the real-world hazy image, and the dehazing results of (b) DCP \citep{he2011single}, (c) AOD-Net \citep{li2017aod}, (d) FD-GAN \citep{dong2020fd}, (e) Semi-dehazing \citep{li2020semi}, (f) PSD \citep{chen2021psd}, and (g) our UCL-Dehaze, respectively. Our UCL-Dehaze can produce both haze-free and more natural images}
	\label{fig:fig6}
\end{figure*}

\subsection{Comparison with State-of-the-arts}
\textbf{Results on Synthetic Dataset.} We report the averaged PSNR and SSIM of 18 state-of-the-art dehazing methods on SOTS outdoor and HSTS datasets in Table \ref{Tab1}. Since haze affects only outdoor vision systems, we mainly focus on evaluating these methods on outdoor datasets. For all compared approaches, we either retrain their models on the ITS dataset or directly use the pre-trained models provided by the authors for evaluation. Obviously, UCL-Dehaze achieves the highest PSNR and SSIM values by a large margin on both datasets, compared to the SOTAs.

In addition to PSNR and SSIM, we also report the averaged CIEDE2000, Contrast gain, ($e, \bar{r}$), and $\sigma$ for a comprehensive evaluation of the different dehazing algorithms from color difference, contrast, visibility, and saturation, respectively. As exhibited in Table \ref{Tab22}, the proposed UCL-Dehaze outperforms other dehazing methods in terms of CIE- DE2000, Contrast gain, and $\sigma$, which indicates that the restored images by our method have more realistic colors and higher contrast. Although compared with DCP and PSD, our UCL-Dehaze is not the best in terms of visibility, it still achieves impressive performance and ranks second among the nine dehazing algorithms. 

Moreover, we exhibit qualitative comparisons of the dehazing results in Fig. \ref{fig:fig4} and Fig. \ref{fig:fig5}. We can observe that DCP cannot achieve satisfactory dehazing results due to the color distortion in sky regions. 
Both AOD-Net and FD-GAN avoid color distortion, however, haze residuals will still happen. The dehazing results of Semi-dehazing are pretty good but still cannot completely remove the haze in some regions. PSD can improve the overall visibility of the hazy images, but the results seem to be very bright and unnatural. 
In contrast, our UCL-Dehaze can produce much clearer and more natural dehazing results. 

\textbf{Results on Real-world Hazy Images.} To evaluate the performance of UCL-Dehaze in real-world hazy conditions, we compare our method with SOTAs on real-world hazy images. 
Fig. \ref{fig:fig6} exhibits five real-world hazy samples and the dehazing results by different approaches. 
Similar to the results in Fig. \ref{fig:fig4} and Fig. \ref{fig:fig5}, DCP cannot dehaze the sky regions well and introduce artifacts. The dehazing results of AOD-Net and FD-GAN still have haze residuals. Semi-dehazing tends to darken the images and cannot remove the haze completely. The results of PSD look very bright and unnatural, and there is still some remaining haze. Compared with these SOTAs, our UCL-Dehaze produces the most natural haze-free images with perceptually pleasing and consistent quality.

Furthermore, we conduct a user study to better understand the performance of UCL-Dehaze on real-world images. In detail, we prepare 50 real-world hazy images randomly selected from the URHI dataset. Then, we adopt five state-of-the-art dehazing methods and our UCL-Dehaze to remove the haze from these 50 images. Next, we recruit 10 participants (5 males and 5 females) and ask them to score the dehazing results on a scale from 1 to 10, where 1 refers to the worst quality and 10 refers to the best quality. For each participant, we present him/her the 300 dehazing images in a random order without showing the corresponding dehazing methods. As demonstrated in Table \ref{Tab:2}, our UCL-Dehaze achieves the best performance for real-world haze removal.

\begin{table}[htbp]
\centering
	\caption{User study. Mean ratings are given ± standard deviation for each approach}
	\label{Tab:2}
	\begin{threeparttable}
    \footnotesize
	\centering
			\begin{tabular}{lc}
				\toprule
				\multirow{1}{*}{Method}&
                \multirow{1}{*}{Rating (mean \& standard dev.)} \cr
				\midrule
				DCP   & 4.43 ± 1.59  \cr
				AOD-Net     & 4.96 ± 1.04  \cr
				FD-GAN   &5.66 ± 1.07 \cr
				Semi-dehazing   & 5.89 ± 0.74  \cr
				PSD  & 6.17 ± 0.89 \cr
				UCL-Dehaze (ours)   & \textbf{6.59} ± \textbf{0.82}  \cr
				\bottomrule
			\end{tabular}
	\end{threeparttable}
\end{table}

\begin{table*}[htbp]
	\centering
	\caption{Quantitative comparisons (NIQE/BRISQUE/SSEQ/PI) with SOTAs on 50 real-world images. \textcolor{red}{Red} and \textcolor{blue}{blue} indicate the $1^{st}$ and $2^{nd}$ ranks, respectively}
	\label{Tab23}
	\begin{threeparttable}
		\footnotesize
		\centering
		\setlength{\tabcolsep}{0.9mm}{
			\begin{tabular}{llcccc}
				\toprule
				\multirow{1}{*}{Method}&
				\multirow{1}{*}{Type}&
				\multirow{1}{*}{NIQE$\downarrow$}&
				\multirow{1}{*}{BRISQUE$\downarrow$}&
				\multirow{1}{*}{SSEQ$\downarrow$}&
				\multirow{1}{*}{PI$\downarrow$}\cr
				\midrule
				Hazy  &  -  & 4.390 & 31.515 & 29.665 & 4.097 \cr
				DCP \citep{he2011single}  &  Prior & \textcolor{blue}{3.743} & 26.831 & \textcolor{blue}{26.146} & 3.418 \cr
				AOD-Net \citep{li2017aod}  &  Supervised & 4.117 & 28.019 & 27.076 & 3.435 \cr
				FD-GAN \citep{dong2020fd}  &  Supervised & 3.825 & 25.886 & 27.718 & 3.463 \cr
				Semi-dehazing \citep{li2020semi}  & Semi-supervised & 4.062 & \textcolor{blue}{24.768} & 29.321 & 3.457 \cr
				PSD \citep{chen2021psd}  & Unsupervised & 3.775 & 25.265 & 28.549 & \textcolor{red}{3.342} \cr
				UCL-Dehaze & Unsupervised & \textcolor{red}{3.736} & \textcolor{red}{24.658} & \textcolor{red}{26.028} & \textcolor{blue}{3.412} \cr
				\bottomrule
			\end{tabular}
		}
	\end{threeparttable}
\end{table*}

For the quantitative comparison, we employ four well-known no-reference image quality assessment metrics: NIQE \citep{DBLP:journals/spl/MittalSB13}, BRISQUE \citep{DBLP:journals/tip/MittalMB12}, SSEQ \citep{DBLP:journals/spic/LiuLHB14}, and PI \citep{DBLP:conf/eccv/BlauMTMZ18}. All these indicators are evaluated on the 50 images prepared for the user study. Evaluation results are illustrated in Table \ref{Tab23}. NIQE and BRISQUE are used to evaluate the overall quality of the images, and lower values indicate better results. As a display, our UCL-Dehaze achieves the best performance in these two indicators. SSEQ evaluates image quality by counting the entropy in the spatial and frequency domains of image patches, and UCL-Dehaze wins first place again, which indicates that the images restored by UCL-Dehaze are clean and perceptually pleasing. PI is a criterion that bridges the visual effect with computable index and has been widely used in the field of image super-resolution. Clearly, our UCL-Dehaze also achieves impressive performance in terms of PI. In general, UCL-Dehaze wins three of the four indicators, which further verifies the superiority of our method on real-world dehazing tasks.

\subsection{Ablation Study} \label{ablation}
\textbf{Effect of different components in UCL-Dehaze.} The proposed network shows superior dehazing performance compared to SOTAs. To further study the effectiveness of UCL-Dehaze, we implement extensive ablation studies to analyze the effectiveness of its components. 

\begin{figure*}[!ht]
	\includegraphics[width=0.9\linewidth]{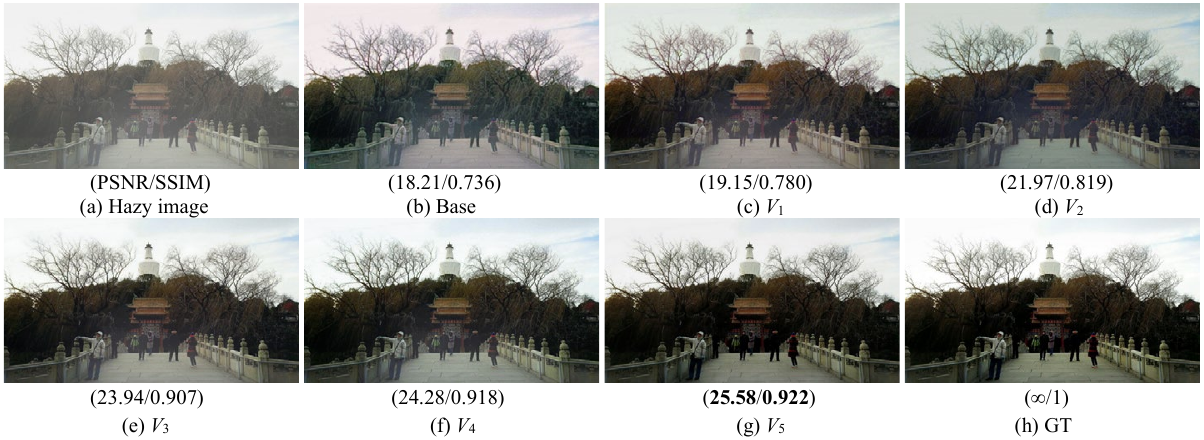}
	\centering
	\caption{Visual results of ablation studies. From (a) to (h): (a) the input hazy image, and the results of (b) Base, (c) $V_1$, (d) $V_2$, (e) $V_3$, (f) $V_4$, (g) $V_5$, respectively, and (h) the ground-truth image.}
	\label{as}
\end{figure*}

We first construct our base network with the original ResNet-based generator, and then we train this model through the LSGAN loss and unidirectional $L_{PC}$. Subsequently, we incrementally add different components into the base network as follows:
\begin{enumerate}
\item  base network + $L_{ide}$ $\rightarrow$ $V_1$, 
\item  $V_1$ + dual-direction $L_{PC}$ $\rightarrow$ $V_2$,
\item  $V_2$ + self-contrastive perceptual loss $L_{SCP}$ $\rightarrow$ $V_3$,
\item  $V_3$ + spectral normalization $\rightarrow$ $V_4$,
\item  $V_4$ + self-calibrated convolutions $\rightarrow$ $V_5$ (full model).
\end{enumerate}
All these variants are retrained in the same way as before and tested on the HSTS dataset. The performances of these variants are summarized in Table \ref{Tab3} and Fig. \ref{as}. 

\begin{table}[htbp]
	\centering
	\caption{Ablation Analysis on UCL-Dehaze. Our full model outperforms its alternatives}
	\label{Tab3}
	\begin{threeparttable}
        \footnotesize
		\centering
		\setlength{\tabcolsep}{1.2mm}{
			\begin{tabular}{lllllll}
				\toprule
				\multirow{1}{*}{Variants}&
				\multirow{1}{*}{Base}&
				\multirow{1}{*}{$V_1$}&
				\multirow{1}{*}{$V_2$}&
				\multirow{1}{*}{$V_3$}&
				\multirow{1}{*}{$V_4$}&
				\multirow{1}{*}{$V_5$} \cr
				\midrule
				
				$L_{ide}$    & w/o & \checkmark & \checkmark & \checkmark & \checkmark & \checkmark \cr
				Dual-$L_{PC}$    & w/o & w/o & \checkmark & \checkmark & \checkmark & \checkmark \cr
				$L_{SCP}$  & w/o & w/o & w/o & \checkmark & \checkmark & \checkmark \cr
				Sp-Norm & w/o & w/o & w/o & w/o & \checkmark & \checkmark \cr
				SC Conv & w/o & w/o & w/o & w/o & w/o & \checkmark \cr
				\midrule
				PSNR & 19.27 & 21.14 & 22.21 & 25.51 & 25.97 & \textbf{26.87}\cr
				SSIM & 0.778 & 0.812 & 0.837 & 0.914 & 0.922 & \textbf{0.933}\cr
				\bottomrule
			\end{tabular}
		}
	\end{threeparttable}
\end{table}

As shown, each component of our UCL-Dehaze contributes to image dehazing, especially the proposed self-contrastive perceptual loss $L_{SCP}$, which achieves 3.3dB PSNR gains over variant $V_2$. If we fully adopt the implementation details in this work, the dehazing results will perform favorably against the state-of-the-art methods.

\textbf{Effect of the weights in loss functions.} To improve the quality of the final restored images and make them much clearer, we exploit comprehensive loss functions that contain adversarial loss, patch-wise contrastive loss, self-contrastive perceptual loss, and identity loss. Accordingly, four loss weights (i.e., $\lambda_{1}$, $\lambda_{2}$, $\lambda_{3}$ and $\lambda_{4}$) are proposed to balance the performance of different loss functions. For all these loss weights, numerous experiments are performed on the HSTS dataset to ensure their optimum values, as shown in Table \ref{Tab:4}, Table \ref{Tab:5} and Table \ref{Tab:6}. As displayed, the validation of the experiments further demonstrates the rationality of the hyperparameter settings. Therefore, when setting $\lambda_{1}=1$ (following CUT \citep{park2020contrastive} and CWR \citep{han2021underwater}), $\lambda_{2}=1$, $\lambda_{3}=0.0002$ and $\lambda_{4}=5$ in our experiments, the performance of UCL-Dehaze is the best.

\begin{table}[htbp]
	\centering
	\footnotesize
	\caption{Ablation study on the dual-direction patch-wise contrastive loss $L_{PC}$ (hyperparameter $\lambda_{2}$)}
	\label{Tab:4}
	\begin{threeparttable}
		\centering
		\setlength{\tabcolsep}{1.2mm}{
			\begin{tabular}{lcccccc}
				\toprule
				\multirow{1}{*}{}&
				\multirow{1}{*}{$\lambda_{2}=0.1$}&
				\multirow{1}{*}{$\lambda_{2}=0.5$}&
				\multirow{1}{*}{$\lambda_{2}=1$}&
				\multirow{1}{*}{$\lambda_{2}=2$}\cr
				\midrule
				PSNR & 26.15 & 26.37 & \textbf{26.87} & 25.42\cr
				SSIM & 0.899 & 0.907 & \textbf{0.933} & 0.865\cr
				\bottomrule
			\end{tabular}
		}
	\end{threeparttable}
\end{table}

\begin{table}[htbp]
	\centering
	\footnotesize
	\caption{Ablation study on the self-contrastive perceptual loss $L_{SCP}$ (hyperparameter $\lambda_{3}$)}
	\label{Tab:5}
	\begin{threeparttable}
		\centering
		\setlength{\tabcolsep}{0.8mm}{
			\begin{tabular}{lcccccc}
				\toprule
				\multirow{1}{*}{}&
				\multirow{1}{*}{$\lambda_{3}=0.00005$}&
				\multirow{1}{*}{$\lambda_{3}=0.0001$}&
				\multirow{1}{*}{$\lambda_{3}=0.0002$}&
				\multirow{1}{*}{$\lambda_{3}=0.0005$}\cr
				\midrule
				PSNR & 25.57 & 26.05 & \textbf{26.87} & 26.27 \cr
				SSIM & 0.917 & 0.925 & \textbf{0.933} & 0.906 \cr
				\bottomrule
			\end{tabular}
		}
	\end{threeparttable}
\end{table}

\begin{table}[htbp]
	\centering
	\footnotesize
	\caption{Ablation study on the identity loss $L_{ide}$ (hyperparameter $\lambda_{4}$)}
	\label{Tab:6}
	\begin{threeparttable}
		\centering
		\setlength{\tabcolsep}{1.2mm}{
			\begin{tabular}{lcccccc}
				\toprule
				\multirow{1}{*}{}&
				\multirow{1}{*}{$\lambda_{4}=1$}&
				\multirow{1}{*}{$\lambda_{4}=2$}&
				\multirow{1}{*}{$\lambda_{4}=5$}&
				\multirow{1}{*}{$\lambda_{4}=10$}\cr
				\midrule
				PSNR & 25.78 & 26.29 & \textbf{26.87} & 26.35 \cr
				SSIM & 0.903 & 0.910 & \textbf{0.933} & 0.920 \cr
				\bottomrule
			\end{tabular}
		}
	\end{threeparttable}
\end{table}

\subsection{Runtime Analysis}
Efficiency is essential for a computer vision system \citep{pang2016cascade}. We evaluate the computational performance of various state-of-the-art dehazing methods and report their average running times in Table \ref{Tab:3}. All the approaches are implemented on a system with an Intel(R) Core(TM) i9-10920X CPU, 32 GB RAM, and an NVIDIA GeForce RTX 3090 GPU. It can be seen that the proposed UCL-Dehaze takes about 0.08$s$ to process one hazy image from the HSTS dataset on average, which is faster and more efficient than other methods.

\begin{table}[htbp]
\centering
	\caption{Average running times (seconds) of different methods tested on the HSTS dataset}
	\label{Tab:3}
	\begin{threeparttable}
    \footnotesize
	\centering
			\begin{tabular}{lll}
				\toprule
				\multirow{1}{*}{Method}&
                \multirow{1}{*}{Platform} &
                \multirow{1}{*}{Average time} \cr
				\midrule
				DCP    & Python (CPU) & 1.41  \cr
				AOD-Net      & PyTorch (GPU) & 0.11  \cr
				GFN   & PyTorch (GPU) & 0.44 \cr
				Cycle-Dehaze  & TensorFlow (GPU) & 1.97 \cr
				GCANet  & PyTorch (GPU) & 0.21 \cr
				Deep DCP   & TensorFlow (GPU) & 0.65  \cr
				PSD   & PyTorch (GPU) & 0.39 \cr
				UCL-Dehaze (ours)   & PyTorch (GPU) & \textbf{0.08}  \cr
				\bottomrule
			\end{tabular}
	\end{threeparttable}
\end{table}

\begin{figure*}[!h] \centering
	\includegraphics[width=1.0\linewidth]{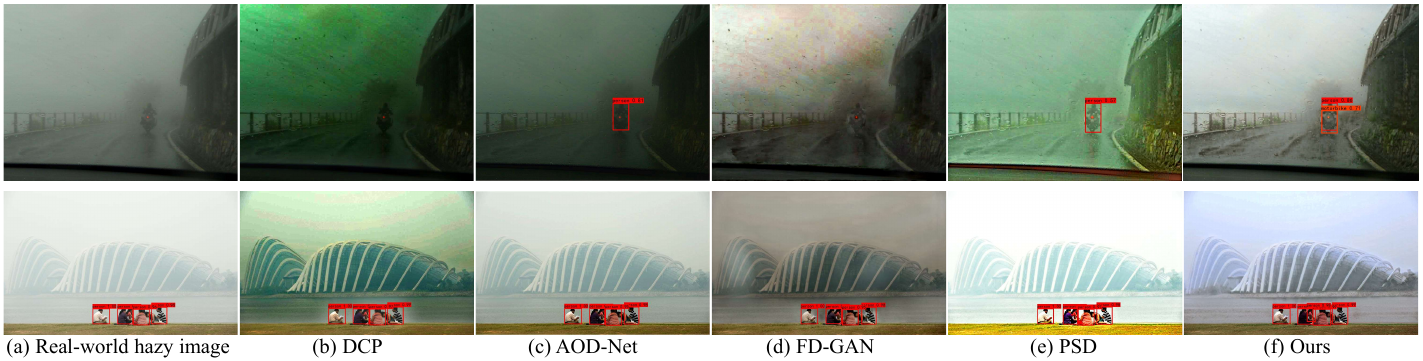}
	\caption{
Object detection results on real-world hazy images and images after dehazing by different approaches. From (a) to (f): (a) object detection results on the hazy images, and the detection results after dehazing by (b) DCP \citep{he2011single}, (c) AOD-Net \citep{li2017aod}, (d) FD-GAN \citep{dong2020fd}, (e) PSD \citep{chen2021psd}, and (f) our UCL-Dehaze, respectively. Please zoom in for best view}
	\label{fig:fig8}
\end{figure*}

\begin{figure*}[htbp] \centering
	\includegraphics[width=0.9\linewidth]{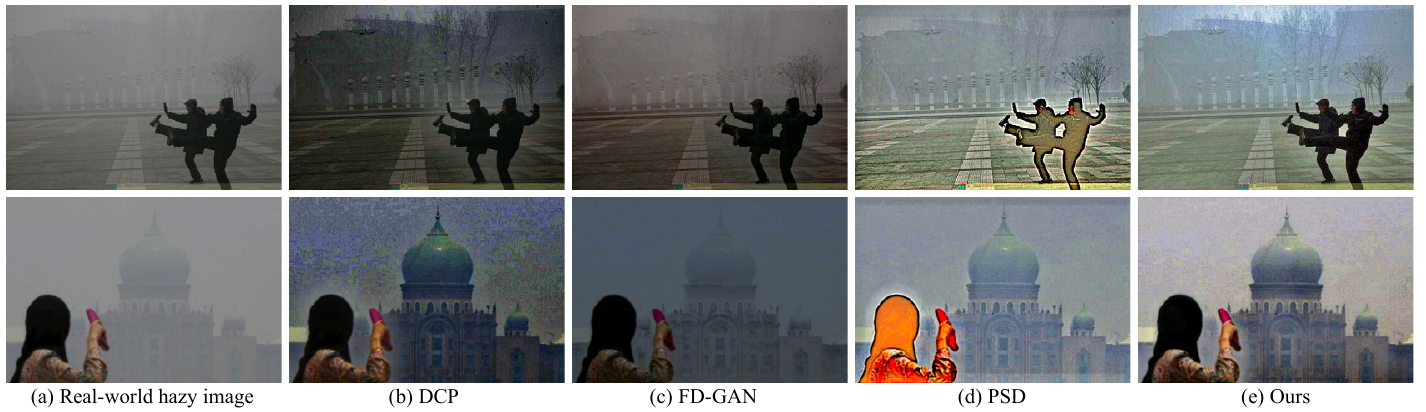}
	\caption{
Typical failure cases of different methods. From (a) to (e): (a) the input hazy image, and the results of (b) DCP \citep{he2011single}, (c) FD-GAN \citep{dong2020fd}, (d) PSD \citep{chen2021psd}, and (e) our UCL-Dehaze, respectively}
	\label{fig8}
\end{figure*}

\subsection{Application}
The performance of object detection algorithms can be significantly degraded when the images are corrupted in hazy conditions. To further demonstrate that the proposed UCL-Dehaze benefits vision-based systems (e.g., traffic monitoring and outdoor surveillance), we employ a pre-trained YO- LOv4 detector \citep{bochkovskiy2020yolov4} to detect objects of interest on real-world hazy images and the corresponding dehazing results by different approaches. As exhibited in Fig. \ref{fig:fig8}, after dehazing, the confidences in detecting objects of interest are greatly improved. Moreover, our UCL-Dehaze outperforms the other dehazing algorithms for object detection on hazy images.

\subsection{Limitations and Discussion}
Although our UCL-Dehaze has achieved compelling results on both synthetic datasets and real-world scenarios, there is still a slight gap between its dehazing performance (on synthetic datasets) and the SOTA supervised-based methods (e.g., DIDH \citep{shyam2021towards}, \citep{wu2021contrastive}). However, due to the existence of domain shifts, the dehazing ability of UCL-Dehaze can surpass these methods in real-world scenes.

Moreover, similar to other dehazing approaches (e.g., DCP \citep{he2011single}, FD-GAN \citep{dong2020fd}, PSD \citep{chen2021psd}, etc.), we find that UCL-Dehaze is not very robust for the heavily hazy scenes. We provide two typical failure cases in Fig. \ref{fig8}. It can be observed that the overall scene and the edges of objects in heavy haze are difficult to recover naturally. In the near future, we will make efforts to solve this limitation.

\section{Conclusion}
In this paper, we avoid bridging the gap between synthetic and real-world haze.
We explore unsupervised contrastive learning from an adversarial training perspective to leverage unpaired real-world hazy and clean images.
Accordingly, we propose an effective unsupervised contrastive learning paradigm for image dehazing, termed UCL-Dehaze.
Unlike most existing image dehazing works, UCL-Dehaze does not require paired data during training and utilizes unpaired positive/negative data to better enhance the dehazing performance. 
It leverages an adversarial training effort and benefits from unpaired real-world training data, thus can generalize smoothly in real-world hazy scenarios. 
In addition, to effectively train the network in an unsupervised manner, we formulate a new pixel-wise contrastive loss function, i.e., the self-contrastive perceptual loss, which encourages the restored images to approach the clean images while keeping away from the hazy ones in the embedding space. 
Finally, comprehensive evaluations demonstrate that our method performs favorably against the state-of-the-arts, even only 1,800 unpaired real-world images are consumed to train our UCL-Dehaze.


{\small
    \bibliographystyle{spbasic}
    \bibliography{ref}
}

\end{document}